\documentclass[journal]{IEEEtran}

\usepackage{epsfig}
\usepackage{graphicx}
\usepackage{amsmath}
\usepackage{amssymb}
\usepackage{bbding}
\usepackage{algorithmicx,algorithm}
\usepackage{algpseudocode}
\usepackage{graphicx}
\usepackage{comment}
\usepackage{url} 
\usepackage{amsmath,amssymb} 
\usepackage{color}
\usepackage{algorithmicx,algorithm}
\usepackage{algpseudocode}
\usepackage{threeparttable}
\usepackage{multirow}
\usepackage{makecell}
\usepackage{booktabs}

\usepackage{epsfig}
\usepackage{bbding}
\usepackage{cite}
\usepackage[T1]{fontenc}

\begin{document}
\title{Person Re-identification via Attention Pyramid}
\author{Guangyi~Chen, Tianpei~Gu,
	Jiwen~Lu,~\IEEEmembership{Senior~Member,~IEEE,}
	Jin-An~Bao,
	and~Jie~Zhou,~\IEEEmembership{Senior~Member,~IEEE} 
	\thanks{\IEEEcompsocthanksitem 
		The authors are with the Beijing National Research Center for Information Science and Technology (BNRist), Department of Automation, Tsinghua University, Beijing, 100084, China. Email: chen-gy16@mails.tsinghua.edu.cn; brucegu@umd.edu; lujiwen@tsinghua.edu.cn; bja1021@bupt.edu.cn; jzhou@tsinghua.edu.cn.
		}} 
\maketitle
\begin{abstract}
In this paper, we propose an attention pyramid method for person re-identification. Unlike conventional attention-based methods which only learn a global attention map, our attention pyramid exploits the attention regions in a multi-scale manner because human attention varies with different scales. Our attention pyramid imitates the process of human visual perception which tends to notice the foreground person over the cluttered background, and further focus on the specific color of the shirt with close observation. Specifically, we describe our attention pyramid by a ``split-attend-merge-stack'' principle. We first split the features into multiple local parts and learn the corresponding attentions. Then, we merge local attentions and stack these merged attentions with the residual connection as an attention pyramid. The proposed attention pyramid is a lightweight plug-and-play module that can be applied to off-the-shelf models. We implement our attention pyramid method in two different attention mechanisms including: channel-wise attention and spatial attention. We evaluate our method on four large-scale person re-identification benchmarks including Market-1501, DukeMTMC, CUHK03, and MSMT17. Experimental results demonstrate the superiority of our method, which outperforms the state-of-the-art methods by a large margin with limited computational cost. $\footnote{Code is available at https://github.com/CHENGY12/APNet}$
\end{abstract}


\IEEEpeerreviewmaketitle

\section{Introduction}

Person Re-Identification (ReID) focuses on matching the images or videos of the same person captured from non-overlapping cameras, which is of paramount importance for many applications, such as suspect tracking and missing person retrieval. It has been significantly advanced in recent years with the aggressive improvement of deep learning. Despite the recent progress, learning a discriminative feature to identify the person from a large set of gallery candidates is still challenging due to large intra-class variance caused by pose variations, occlusions, or cluttered backgrounds.

\begin{figure}[t]
	\centering
	\includegraphics[width=0.99\linewidth]{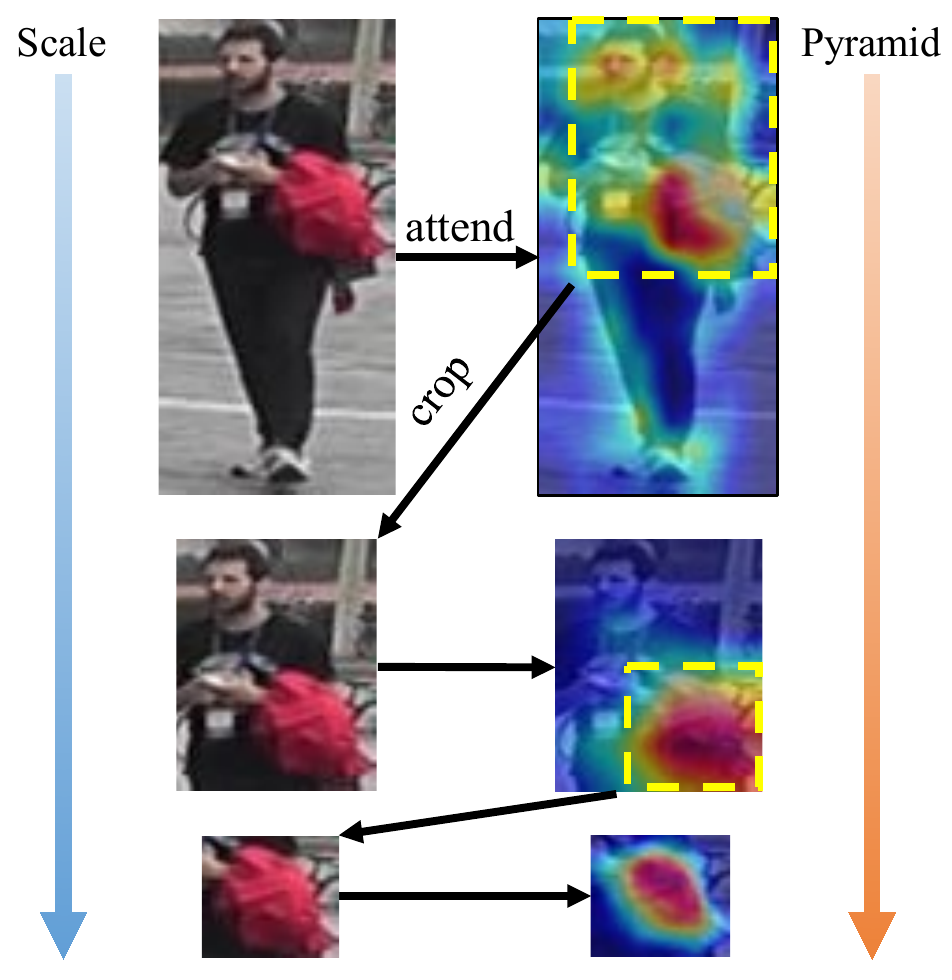}
	\caption{The motivation of the attention pyramid network. The attentions of the human vision system are changing with the visual scale. When we look at a whole image, the upper part of the person attracts our attentions. This attention will further be turned to the red coat in the arms when we have a close look.}
	\label{fig:motivation}
\end{figure}

Recently, attention mechanism\cite{zhang2020relation,chen2019self,chen2019spatial,xu2018attention,li2018harmonious,fang2019bilinear} has been widely used in the ReID system to facilitate high-performance identification and demonstrates the powerful representation ability by discovering discriminative regions and mitigating the misalignment. For example, Zhang \emph{et al.}~\cite{zhang2020relation} proposed a relation-aware attention model to focus on the inter-relation in the feature map. Li \emph{et al.}~\cite{li2018harmonious} introduced harmonious attention to simultaneously learn hard region-level and soft pixel-level attention. 
 
These methods learn to explore salient regions in the global image, which can be formulated as a salient detection task. However, detecting the salient regions with the attention model is confronted with the dilemma to jointly capture both coarse and fine-grained clues, since the focus varies as the image scale changes. As shown in Fig.~\ref{fig:motivation}, we tend to focus on the upper part of the person with the given whole image, transfer our sight to the salient coat in the arms given the upper part, and further concentrate on the more discriminative regions. It is consistent with the human vision system attending salient objects sequentially, from coarse to fine.  

To address the above issue, we propose effective attention pyramid networks (APNet) to jointly learn the attentions under different scales. It is motivated by the widely-used feature pyramid method (FPN)~\cite{lin2017feature} in the visual detection system, which captures multi-scale clues with the pyramid structure. We regard attention learning as a process of salience detection and learn the attention pyramid for multi-scale saliences. 
Unlike the FPN applying the feature maps of high resolution in shallow layers to detect the small objects and low resolution in deep layers to detect large objects, our APNet focuses on learning the discriminative representation covering the salient regions from coarse to fine. Therefore, we propose a ``split-attend-merge-stack'' principle to build our attention pyramid, which splits the feature maps into different granularities and aggregates the attentive clues by stacking the attention module learned from different granularities. In each level of our attention pyramid, we split the features into multiple local parts and learn the attention maps of each part. Then, we merge these local attention maps to obtain the attention of the global image. At different pyramid levels, we split the features into local parts with different granularities to capture the salient clues in multiple scales. Finally, we stack the attention maps from coarse to fine as a pyramid structure to aggregate clues from different granularities. 

Compared with the traditional attention model, our APNet effectively captures the discriminative clues with different scales with the proposed ``split-attend-merge-stack'' principle. 
Compared with other feature pyramid learning methods which extract features of different scales and aggregate them, our APNet requires no extra feature extraction module by replacing it with the splitting and stacking patterns.
Therefore, our APNet can be easily integrated into any baseline attention model with a lightweight computational cost (comparable with attention learning). 
Beyond spatial attention, our attention pyramid framework can be applied to other attention modules, such as channel-wise attention or temporal attention. To evaluate the generality of our attention pyramid framework, we implement our attention pyramid framework for channel-wise attention module and spatial attention module, which respectively explores the discriminative clues in channel-wise and spatial domains. 
We conduct extensive experiments to evaluate our APNet on four popular person re-identification benchmarks including Market-1501~\cite{zheng2015scalable}, DukeMTMC-reID~\cite{ristani2016performance}, CUHK03~\cite{li2014deepreid}and MSMT17~\cite{wei2018person}. 
Experimental results demonstrate that our APNet outperforms the state-of-the-art methods by a large margin with limited computational cost. Besides, we also conduct the cross-dataset evaluation and occlusion evaluation to evaluate the generalization ability and robustness of our method.

We summarize the contributions of this work as follows:

\begin{itemize}
\item [1)] We propose the attention pyramid networks for person ReID, which jointly explores the salient clues in different scales by the proposed ``split-attend-merge-stack'' principle.
\item [2)] We implement our attention pyramid framework for different attention modules including channel-wise and spatial ones.
\item [3)] In the experiments, our method achieves obvious improvement with limited computational cost and we demonstrate better generalization ability and robustness.
\end{itemize}

\section{Related Work}

In this section, we briefly review three related topics, including deep person re-identification, attention mechanism, and feature pyramid.

\subsection{Deep Person Re-identification}
Recent Person ReID methods obtain excellent performance with the development of deep learning models, which learn the robust feature representation for the misaligned person image. Most of these methods are categorized into two research fields including capturing more prior knowledge or supervisory signals and designing more effective networks. To mine more clues as the prior knowledge, some methods utilize body structure knowledge~\cite{Zhao_2017_CVPR,Li_2017_CVPR,zhang2019densely,kalayeh2018human} and human pose information~\cite{Su_2017_ICCV,qian2018pose} for accurate part detection or person normalization. For example, Kalayeh~\emph{et al.}~\cite{kalayeh2018human} propose to use the human semantic parsing as the prior knowledge to refine the person ReID model. While PN-GAN~\cite{kalayeh2018human} estimates the human pose and normalizes it by a generative adversarial network to mitigate the influence of pose variations. Besides, attribute labels~\cite{zhao2019attribute,tay2019aanet} and spatial-temporal pattern~\cite{wang2019spatial,lv2018unsupervised} are also introduced in the Person ReID system as the complementary supervisory signal to improve performance. AAnet~\cite{tay2019aanet} integrates person attributes and attribute attention maps into a unified learning framework to guide attention learning with the attribute labels. St-ReID~\cite{wang2019spatial} utilizes the spatial-temporal information in the camera network to filter irrelevant negative samples of gallery set and significantly improve the performance. Furthermore, some methods~\cite{hermans2017defense,zheng2019pyramidal,Chen_2017_CVPR,sun2020circle,yao2019deep} aim to optimize the loss function to mine the relations of instances and learn the discriminative embeddings. For example, TriNet~\cite{hermans2017defense} first proposes to introduce the triplet loss into the deep ReID system and achieves excellent performance. While Chen~\emph{et al.}~\cite{Chen_2017_CVPR} further improve it by a margin-based online hard negative mining strategy to enhance the generalization ability of the ReID model. Network designing is another important direction of person ReID. In the early research of deep person ReID, many methods are proposed to explore effective network structure such as earliest Deepreid~\cite{li2014deepreid}, region-based SpindleNet~\cite{Zhao_2017_CVPR}, effective OSNet~\cite{zhou2019omni} and the strong baseline BOT~\cite{luo2019bag}. Recently, many methods focus on designing the part-based model~\cite{sun2018beyond,cheng2016person,chen2019spatial,wang2018learning,Sun_2019_CVPR}, which split the feature maps into multiple parts to learn local features and aggregate them for recognition. Despite significant performance improvement, these methods always suffer the high computational cost. Attention model~\cite{si2018dual,chen2019self,li2018harmonious,chen2019abd,Xia_2019_ICCV,Fang_2019_ICCV,Chen_2019_ICCV2,zhou2019discriminative,martinel2020deep} is also an important direction to design novel network architecture, which discovers salient regions and mitigates the misalignment to learn robust representation. Furthermore, some methods~\cite{quan2019auto} are proposed to automatically search the network architectures for person ReID task.

\subsection{Attention Model}
Attention model~\cite{mnih2014recurrent} naturally imitates the perception of humans to concentrate on what we are interested in. Recently, it has gained great success in many fields, such as visual understanding~\cite{mnih2014recurrent,wang2018non,zhou2018temporal}, natural language processing~\cite{vaswani2017attention}, and graph learning~\cite{velivckovic2017graph}. The attention model also plays an irreplaceable role for the person ReID system to learn discriminative representation. Liu~\emph{et al.}~\cite{liu2017end} introduce the attention model to locate the discriminative salient regions and model it with an RNN. Furthermore, Zhao~\emph{et al.}~\cite{zhao2017deeply} and Xu~\emph{et al.}~\cite{xu2018attention} apply the body part detector to employ the clues of the human body structure in the attention model. Beyond spatial attention, many methods adopt attention mechanisms on the temporal domain~\cite{Liu_2017_CVPR,si2018dual,li2018diversity,zhang2019scan,chen2020learning,chen2020temporal} to explore key temporal frames. The attention model is also applied on the channel domain~\cite{hu2018squeeze,chen2019self} to discover key feature channels, and even on the instance level~\cite{shen2018person,chen2020deep} to capture more valuable instances. Despite the widespread use and the convincing performance of the attention model, the problem of how to jointly capture the salient clues of different scales is still barely studied. SCSN~\cite{chen2020salience} cascades multiple attention models on the extracted features to capture different clues. However, the cascaded structure requires a complex mechanism to avoid information duplication, which is computing expensive for the attention model. In this paper, we focus on designing a basic attention norm to jointly discover the salient clues from coarse to fine, called attention pyramid networks (APNet). Compared with other attention models, the proposed APNet can achieve obvious improvement by the pyramid-structure perception with limited computational cost.

\subsection{Feature Pyramid}
Feature pyramid is a widely-used method to learn multi-scale feature representation for detecting objects of various scales. FPN~\cite{lin2017feature} proposes a top-down pathway to fuse the features with different resolutions and scales. In addition, many variants of FPN are proposed to improve the information propagation ability of FPN, such as PANet~\cite{liu2018path} adds a bottom-up path, Bi-FPN~\cite{tan2020efficientdet} proposes a new cross-scale connection, or connects high-level and low-level features in a nonlinear way~\cite{kong2018deep}. Recently, feature pyramid methods have demonstrated great success in various fields, such as semantic segmentation~\cite{li2018pyramid} and person ReID~\cite{zheng2019pyramidal}. 
Different from the FPN-based methods in the detection task connecting the features of different resolutions, these methods first extract the features with different scales by an extra multi-scale branch and learn to aggregate them. For example, Li~\emph{et al.}~\cite{li2018pyramid} extract features of different scales with multiple convolution blocks and fuse them to learn the attention, while Zheng~\emph{et al.}~\cite{zheng2019pyramidal} split the feature maps into parts with different scales and learn the multi-branch local features of each part. Despite the convincing performance, the extra multi-scale feature representation branch is computationally expensive. In this work, we propose a novel feature pyramid framework with the ``split-attend-merge-stack'' principle, which replaces the complex extra feature extraction module with the splitting and stacking pattern. Thanks to such principle, we learn the pyramid feature with similar cost as the attention learning. Our APNet sequentially learns the attention from coarse to fine, instead of aggregating simultaneously learned features of all scales or parts.

\begin{figure*}[t]
	\centering
	\includegraphics[width=0.99\linewidth]{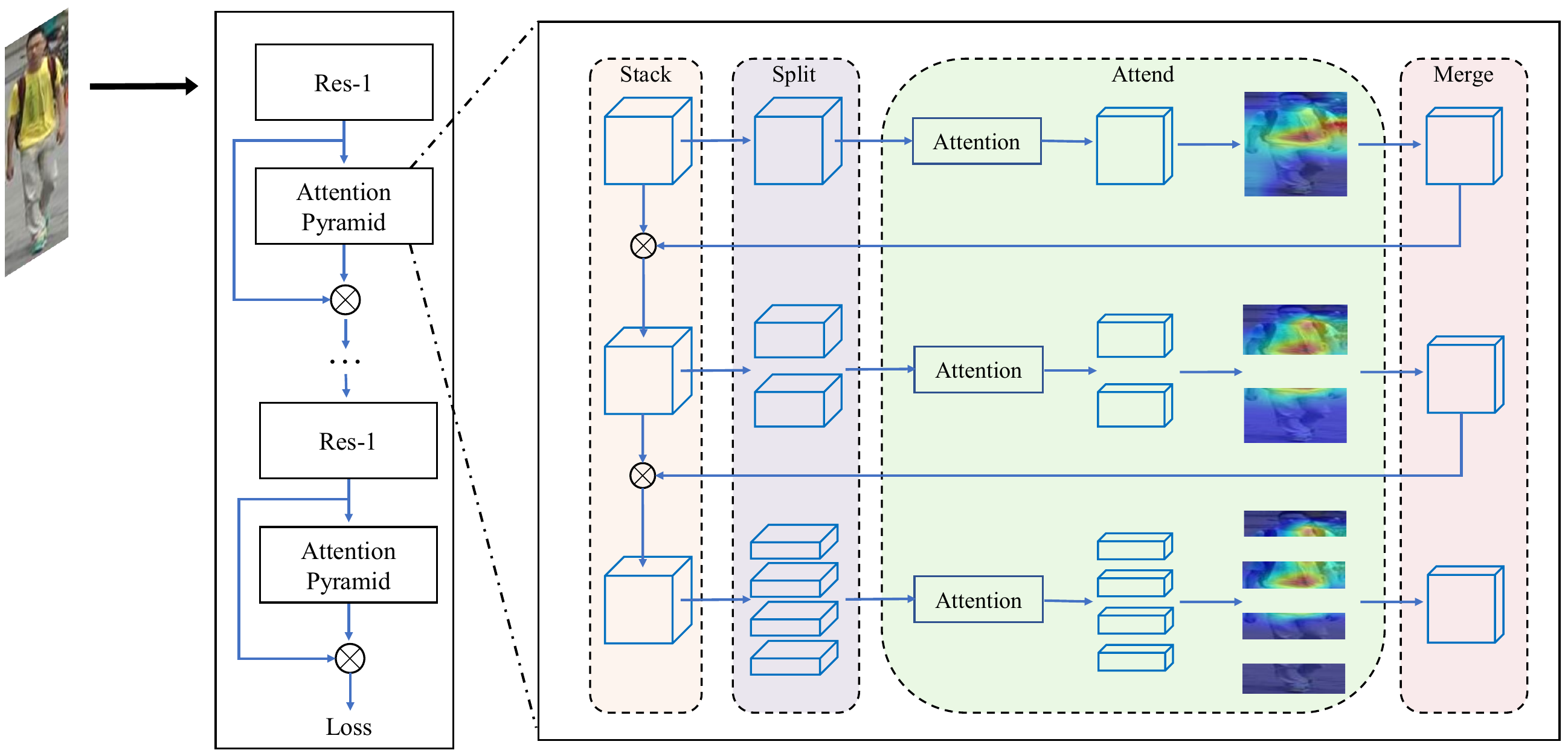}
	\caption{The architecture of Attention Pyramid Networks (APNet). Our APNet adopts the ``split-attend-merge-stack'' principle, which first splits the feature maps into multiple parts, obtains the attention map of each part, and the attention map for the current pyramid level is constructed by merging each attention map. Then in a deeper pyramid level, we split the features into more fine-grained parts and learn the fine-grained attention guiding by coarse attentions. Finally, attentions with different granularities are stacked as an attention pyramid and applied to the original input feature by element-wise product.}
	\label{fig:method}
\end{figure*}

\section{Approach}

In this section, we first introduce our attention pyramid networks (APNet) and implement them on both channel-wise and spatial attention methods. Then, optimization procedure and implementation details will be presented. Finally, we discuss different pyramid networks and explain why the proposed APNet is more effective and efficient for person ReID task.

\subsection{Attention Pyramid}
\label{Attention Pyramid}
Attention mechanism has been proved to be efficient in exploiting discriminative features of the image. However, how to jointly capture the salient clues of different scales with a limited cost is still challenging. To address this problem, we propose attention pyramid networks (APNet) to guide the network sequentially discover salient clues of different scales from coarse to fine, for the comprehensive and complementary perception. Different from conventional feature pyramid methods which extract features with different scales and aggregate them, our APNet captures the multi-scale clues with the attention model. We first learn the coarse attention and use it to guide the more fine-grained attention learning of the split feature maps. We name it as the ``split-attend-merge-stack'' principle in the proposed attention pyramid. In the following, we will introduce this principle in detail and explain why it is effective.

Given a person image $I$, we first extract the feature map $X \in \mathbb{R}^{C \times H \times W}$ by a backbone network block
$\mathcal{F}: I \rightarrow X$, where $C $ denotes the channel dimension and $ H \times W $ denotes the spatial domain. Then, we learn the attention pyramid $\mathcal{P}$ to discover multi-scale salient parts on the feature $X$. The attention pyramid contains attention maps with different levels as $\mathcal{P} = \{ \mathcal{P}_{i} \}  $, where $\mathcal{P}_{i}$ denotes the $i$th pyramid layer. In different pyramid layers, we learn the discriminative clues of different scales. The coarse attentions guide the following fine-grained attention learning. It can be formulated as a sequential learning process as:
\begin{equation}
\begin{aligned}
\label{eq: pyramid}  {X}_{i} = \mathcal{P}_{i}({X}_{i-1}),
\end{aligned}
\end{equation}
where $X_i$ denotes the feature map in the $i_{th}$ level. As shown in Fig.~\ref{fig:method}, for the attention learning of each level, we first split it into more fine-grained granularities and merge these learned attentions as the attention map of the current level. 

\textbf{Split:} The split operation takes the feature map $X_{i}$ as input and outputs $n $ split feature tensors $X_{i,j}$, where $j = \{1,2,...,n\} $ is the index of split feature parts. Specifically, the number of split parts is exponential growth based on a radix, such as $n = 2^{i} $ when pyramid level is $i$ and radix is $2$. With the increase of pyramid level $i$, the number of split parts is accordingly increasing which indicates the granularities of these feature parts are more fine-grained. The split operation is applied to the corresponding domain of the attention modules. For example, we tend to slice the feature map $X_{i}$ into multiple parts $X_{i,j} \in \mathbb{R}^{C \times \frac{H}{n} \times W}$ along the height dimension $H$ for the spatial attention module and apply the division on the channel domain $C$ to obtain $X_{i,j}  \in \mathbb{R}^{\frac{C}{n} \times H \times W}$ for channel-wise attention.

\textbf{Attend $\&$ Merge:} Given split feature tensors $ \{X_{i,j}\} $, we learn a set of sub-attention model $ \{ \mathcal{A}_{i,j} \}  $ to capture the discriminative clues of each feature tensor. Then we merge all sub-attention models as the inverse process of the split operation to obtain the whole attention map $A_{i} $ in the same size, which is formulated as:
\begin{equation}
\begin{aligned}
\label{eq: attend}  A_{i} = [\mathcal{A}_{i,j}(X_{i,j})]_{j=1}^{n},
\end{aligned}
\end{equation}
where $ [\cdot]_{j=1}^{n} $ refers to the concentration of $n$ attention maps. This aggregated attention map $A_{i}$ indicates the salient regions in $i$th level.

\textbf{Stack:} We guide the network to focus on the significant features progressively by stacking attention from coarse to fine as a pyramid. With the learned attention $A_{i}$, the network discovers more discriminative features as:
\begin{equation}
\begin{aligned}
\label{eq: attend2}  X_{i} = \sigma(A_{i}) * X_{i-1},
\end{aligned}
\end{equation}
where $*$ denotes the element-wise product. The features are reweighted with the normalized attention map, and fed into the next pyramid level to guide the more fine-grained attention learning. With the increase of pyramid level, the more fine-grained clues are discovered based on original coarse features. 

\textbf{Multi-Stage Operation:}  
Despite capturing multi-scale discriminative clues with the pyramid structure, it is still challenging to retrieve all salient features at a single step. To alleviate such issue, we apply the proposed attention pyramid at multiple convolution stages of the backbone network, which gradually guides the deep network to discover the salient clues. Specifically, we apply four attention pyramids on the bottom of each residual block of the backbone ResNet50~\cite{he2016deep} network. The multi-stage structure encourages the network to learn more discriminative representation by progressive refinement.

\begin{figure*}[t]
	\centering
	\includegraphics[width=0.99\linewidth]{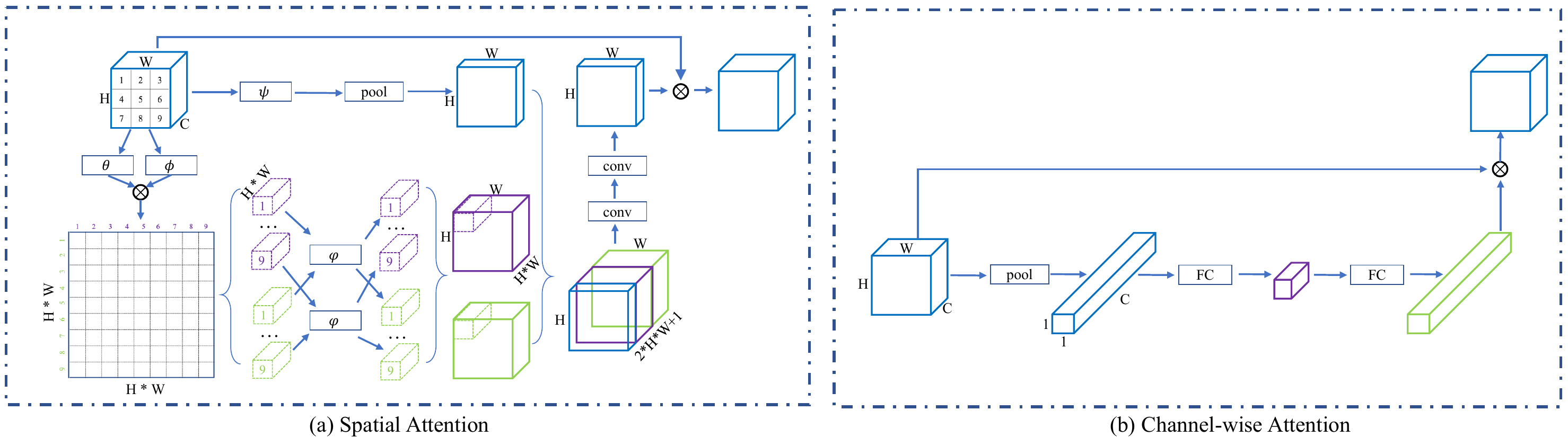}
	\caption{The different basic attention modules of Attention Pyramid Networks (APNet), including spatial attention and channel-wise attention. For spatial attention, we follow the architecture of RGA-S~\cite{zhang2020relation} which models the spatial relations. For channel-wise attention, we adopt the widely used SE attention~\cite{hu2018squeeze,chen2019self} module.}
	\label{fig:attention}
\end{figure*}

\subsection{Attention Model}
\label{Attention Model}
In this work, we implement our attention pyramid framework into two different attention models including channel-wise attention and spatial attention.

\textbf{Spatial Attention:}
Spatial attention focuses the most discriminative region of the input feature map and ignores the irrelevant region (e.g. occlusion). To highlight the contribution of the proposed attention pyramid structure directly, we adopt the relation-aware attention block (RGA-S)~\cite{zhang2020relation} \footnote{we use the official implementation of RGA-S at \url{https://github.com/microsoft/Relation-Aware-Global-Attention-Networks}} as the baseline attention model. Specifically, given a feature tensor ${X} \in \mathbb{R}^{H \times W \times C}$ from a CNN layer, we will learn a spatial attention feature map of size $H \times W$. In this paper, we only simply introduce the basic learning process of the RGA module. The detail architecture and parameter settings can be found in~\cite{zhang2020relation}. As shown in part (a) of Fig.~\ref{fig:attention}, RGA-S first splits $X$ into a set of feature nodes $  \{f_p\}_{p=1:H\times W} | f_{p} \in \mathbb{R}^{C}$. Then the pairwise relations $r_{p,q}$ between feature node $f_{p}$ and feature node $f_{q}$ is calculated as a dot-project affinity as $r_{p,q} = \theta(f_{p})^{T}\phi(f_{q}) $, where $\theta(\cdot)$ and $\phi(\cdot)$ are two $1 \times 1$ convolution layers. These relations help to mine semantics and thus attention. Relation vectors including horizontal $\textbf{r}_{p,:}$ and vertical $\textbf{r}_{:,q}$ are feed into $1 \times 1$ convolution layers and connected as two relation-based attention maps $A^S_{RH} \in \mathbb{R}^{H\times W \times C'} $ and 
$A^S_{RV} \in \mathbb{R}^{H\times W \times C'} $, where $C'=H\times W$ is the channel dimension of attention maps. 
Finally, the spatial attention $ A^{S} $ is learned as 
\begin{equation}
\begin{aligned}
\label{eq: spatial} A^{S} = \sigma(W_{2}^{s}ReLU(W_{1}^{s}[A^S_{RH},A^S_{RV},A^S_G])),
\end{aligned}
\end{equation}
where $W_{1}^{s}$ and $W_{2}^{s}$ are two $1 \times 1$ convolution layers, $[.,.,.]$ refers to feature connection and $A^S_G $ denotes the original global spatial attention which learned by a $1 \times 1$ convolution layer and a pool layer.

\textbf{Channel-wise Attention:}
Channel-wise attention helps the model to focus on the more salient feature by assigning larger weight to channels that show a higher response. We adopt the "Squeeze and Excitation" block (SE Layer)\cite{hu2018squeeze} as the channel-wise attention, which consists of one global average pooling layer and two consecutive fully-connected layers. As shown in part (b) of Fig.~\ref{fig:attention}, given the feature map $X$, the channel-wise attention $A^{C}$ is defined as:  
\begin{equation}
\begin{aligned}
\label{eq: linear} A^{C} = \sigma(W_{2}^{c}ReLU(W_{1}^{c}Pool_{avg}(X)))
\end{aligned}
\end{equation}
where $W_{1}^{c}$ and $W_{2}^{c}$ are the parameters of two fully-connected layers, and $Pool_{avg}$ refers to the average pooling layer. Similar to spatial attention, the final channel-wise attention $A^{C}$ is applied on the original feature map $X$ by channel-wise multiplication. In our attention pyramid framework, we replace the original feature map $X$ with the split sub-features $X_{i,j}$ to learn the local channel-wise sub-attentions and merge them as Equation~\ref{eq: attend}.

\subsection{Optimization}

In the training stage, we use triplet loss and classification loss as identity supervisory to train our APNet following the settings in BOT~\cite{luo2019bag}. The triplet loss optimizes the embedding space to increase the inter-class distance and reduce the intra-class distance. The triplet loss is formulated as:
\begin{equation}
\begin{aligned}
\label{eq: loss_tri} L_{tri}= \frac{1}{N}\sum_{i=1}^{N}\big[|| f_i- f_i^+||-|| f_i- f_i^-||+m \big]_+,
\end{aligned}
\end{equation}
where $[ \cdot]_+$ indicates the max function $ \max(0,\cdot)$, and $f_i,f_i^+,f_i^-$ denote the embedding of the anchor sample, the positive sample, and the negative sample in a batch, respectively. We adopt the hard example mining strategy in our loss function and set the distance margin $m=0.3$ of positive samples to the anchor sample than negative ones. We use the Euclidean distance as the distance metric to learn the triplet loss.
The classification loss is formulated by calculating the cross-entropy between the identity ground-truth and the predicted probability:
\begin{equation}
\begin{aligned}
\label{eq: loss_ce} L_{cls}=  -\frac{1}{N}\sum_{i=1}^{N}{\sum_{k=1}^{K}y_{i,k}\log(p_{i,k})},   
\end{aligned}
\end{equation}
where $y_{i,k} $ denotes the ground-truth label whether the identity of $i$th image is $k$. To overcome the overfitting problem in the training, we transform the above loss function with a label smooth regularization~\cite{szegedy2016rethinking}. In practice, we use a uniform distribution $\mu(k) = 1/K $ to balance the predicted probability as:
\begin{equation}
\begin{aligned}
\label{eq: loss_ce+lsr} L_{cls+lsr}= -\frac{1}{N}\sum_{i=1}^{N}{\sum_{k=1}^{K} \log(p_{i,k}) \big((1-\epsilon)y_{i,k}+\frac{\epsilon}{K} \big) }, 
\end{aligned}
\end{equation}
where the $ 0 < \epsilon < 1 $ is the smoothing rate. The final loss function for the training stage is thus formulated as:
\begin{equation}
\begin{aligned}
\label{eq: objective} L_{apn}=  L_{cls+lsr} + \lambda L_{tri},
\end{aligned}
\end{equation}
where $\lambda$ is the balance rate of two loss functions.

\begin{figure*}[t]
	\centering
	\includegraphics[width=1.0\linewidth]{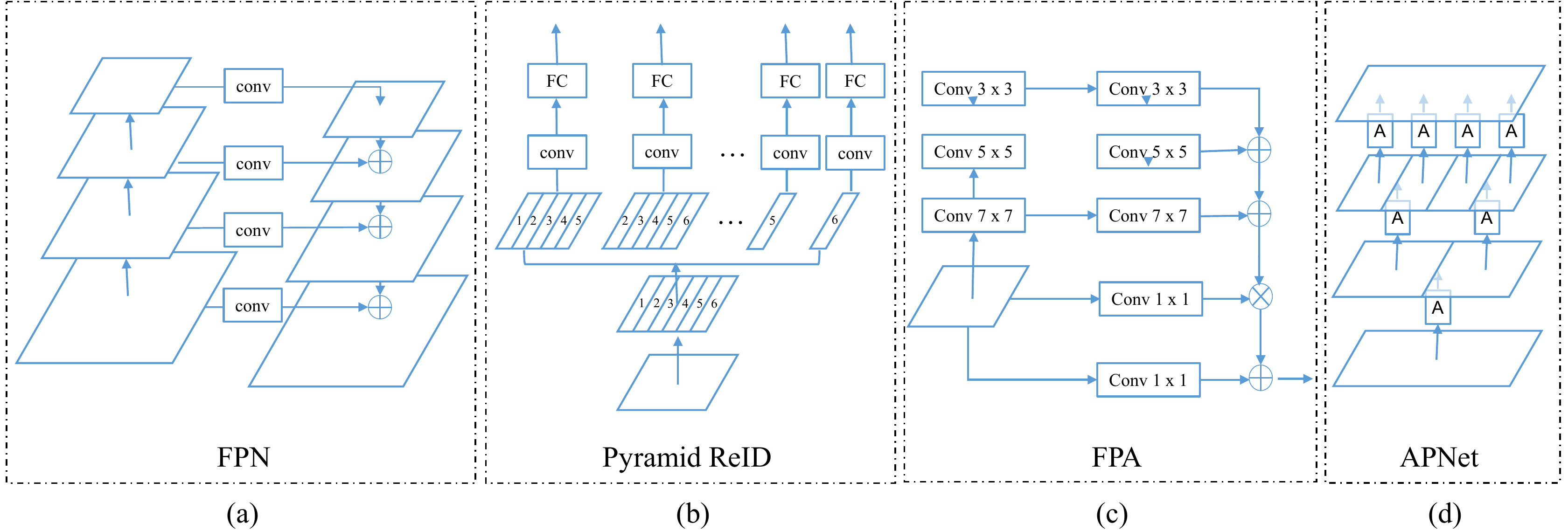}
	\caption{The comparison between different feature pyramid structures. (a) FPN~\cite{lin2017feature} applies a top-down pathway to aggregate features with different resolutions to detect objects with different scales. (b) Pyramid ReID~\cite{zheng2019pyramidal} splits the feature maps with different scales and extracts local features of each part by convolution and fully-connected layers. (c) Feature pyramid attention (FPA)~\cite{li2018pyramid} applies the feature pyramid in the attention model for semantic segmentation by aggregating clues captured with multi-scale convolution layers. (d) Our attention pyramid network proposes a ``split-attend-merge-stack'' principle to learn multi-scale attentions by feature splitting and stacking. It is effective with the coarse-to-fine attention learning and more efficient due to no extra computational cost for feature extraction.}
	\label{fig:discussion}
\end{figure*}

\subsection{Discussion}
In this subsection, we discuss the proposed attention pyramid network with other feature pyramid structures and explain why our APNet is effective and efficient. As shown in Fig.~\ref{fig:discussion}, we compare three different pyramid structures including FPN~\cite{lin2017feature}, Pyramid ReID~\cite{zheng2019pyramidal}, Feature pyramid attention~\cite{li2018pyramid} with our APNet. 

FPN~\cite{lin2017feature} is a widely-used method in the field of object detection, which applies a top-down pathway to aggregate features with different resolutions and scales. However, it is not trivial to directly transfer this structure for the person ReID task. First, FPN aims to balance the resolution and semantic abstraction, which detects small objects with the features in high-resolution and detects large objects with the features in the high semantic information level (low-resolution). While the person ReID task focuses on the identity information behind the images. Thus, the semantic information is of more importance than the resolution for the ReID feature representations. Second, the top-down pathway in the FPN is an extra top-down branch including the convolution layer and upsampling with expensive computational cost. Compared with FPN, our APNet reduces the top-down branch and adds the pyramid structure in the backbone network as the attention model. 

Pyramid ReID~\cite{zheng2019pyramidal} splits the feature map into different parts and learns the local clues of each part with a network branch. However, the computational cost of simultaneously extracting features of all parts is very expensive. While the relation among representations with different scales is not explicitly modeled. In APNet, we reduce the redundant computing by stacking multiple attention layers, where fine-grained attention is guided by the above coarse attention maps. Besides, we show this coarse-to-fine attention model is effective to discover discriminative clues by imitating the human perception process. Some methods apply the same pyramidal structures of Pyramid ReID~\cite{zheng2019pyramidal}, e.g. DPD~\cite{martinel2020deep} method applies multiple pyramidal structures in each stage of backbone. Then, DPD learns an attention model to weight different pyramidal features. 
There are four main differences between APNet and DPD. First, the pyramidal structure and the attention model in DPD are separated. DPD uses the attention to fuse multiple pyramidal features, while APNet aims to learn pyramidal attention to mine multi-scale features. Second, our pyramid structure follows a ``split-attend-merge-stack'' principle, where the multi-scale features are connected with the stacked attention. In DPD, the features with different scales are treated equally, while in APNet, we use the coarse attention to guide the training of the fine-grained ones. Third, to obtain the pyramidal features, APNet applies the attention model while DPD uses a pooling layer. However, the pooling model may reduce the clues. Fourth, for the features in one stage (a block in the backbone network), our APNet applies it as the input of the next stage, while DPD connects all features as the final representation. 

Feature pyramid attention(FPA)~\cite{li2018pyramid} introduces the pyramid structure in the attention model for semantic segmentation problem, which applies convolution layers with multi-scale kernels to extract clues and fuses them to learn the attention model, as shown in Fig.~\ref{fig:discussion} (c). However, there are three problems in the pyramid attention network~\cite{li2018pyramid}. First, it is computing expensive since each attention module consists of multiple multi-scale convolution layers. Second, multi-scale convolution layers only capture the multi-scale information in the spatial domain, which is difficult to transfer for other attention models. Third, feature pyramid attention~\cite{li2018pyramid} fuses the multi-scale information to learn the attention, but not multi-scale attentions. Different from FPA~\cite{li2018pyramid}, our APNet applies the ``split-attend-merge-stack'' principle to learn multi-scale attentions, which captures multi-scale information with multi-granularity split instead of convolution layer. APNet is more efficient due to zero extra computational cost and it can be easily integrated into different attention modules.

\section{Experiments}
We evaluated our method on four large-scale ReID benchmarks including: Market-1501\cite{zheng2015scalable}, DukeMTMC-reID\cite{ristani2016performance}, CUHK03~\cite{li2014deepreid}, and MSMT17\cite{wei2018person}. We conduct extensive ablation studies to investigate the effectiveness of each component in our method and compared our method with other state-of-the-art methods. Besides, we also evaluated the robustness of our method on the occluded person ReID dataset Occluded-DukeMTMC-reID\cite{miao2019PGFA}, and the generalization ability by the cross-dataset experiments.

\begin{table}[!t]
\centering
\normalsize
\caption{The detailed information of datasets.}
\label{tab:info_datasets}
\begin{tabular}{c|c|c|c|c|c}
    \hline
    \multirow{2}*{\textbf{Datasets}}&
    \multicolumn{2}{c|}{Train sets}&\multicolumn{2}{c|}{Test sets}&\multirow{2}*{Cam}\cr\cline{2-5}
    &IDs&images&IDs&images& \cr
    \hline
    Market&751&12936&750&19732&6\cr
    \hline
    Duke&702&16522&702&19889&8\cr
    \hline
    CUHK03&767&7365&700&5332&6\cr
    \hline
    MSMT&4101&32621&3060&93820&15\cr
    \hline
    Occluded-Duke&702&15618&519&19871&8\cr
    \hline
\end{tabular}
\end{table}

\begin{table*}[t]
 \centering
 \normalsize
 \renewcommand\tabcolsep{6pt}
 \caption{Ablation studies of the APNet-C method on the Market-1051, DukeMTMC-reID and MSMT17 dataset.}
 \label{tab: ablation}
 \small
 \begin{tabular}{c c c c c c c c c c}
  \hline
  \multirow{2}{*}{\bf Model}&
  \multicolumn{3}{c}{\bf Market-1501}&
  \multicolumn{3}{c}{\bf DukeMTMC-reID}&
  \multicolumn{3}{c}{\bf MSMT17}\\
  \cline{2-10}
   & \bf mAP &\bf R-1 &\bf R-5& \bf mAP & \bf R-1&\bf R-5 & \bf mAP & \bf R-1&\bf R-5 \\
  \hline
  APN$_{0}$ &87.8& 95.0 & 98.6 & 77.2 & 88.4& 94.4 & 51.2 & 75.3 &84.9  \\
  SE-ResNet & 88.6&95.5 & 98.5& 78.7&88.5& 95.1&58.5& 81.1 &90.3 \\
  \hline
  Stacked Attention (C) & 89.4 &95.3 & 98.7 & 78.7 & 88.4 & 95.1 &58.0&80.5 & 89.5\\
  APNet-C$_{1}$ & 89.6 & 95.5 & 98.6 &79.0 & 88.6 & 95.2 &59.5&  81.9 & 90.7 \\
  APNet-C$_{2}$ & \bf 90.5 & \bf 96.2 & \bf 98.8 & \bf 81.5 &\bf 90.4 & 95.6& \bf 63.5 &\bf 83.7 & \bf 91.7\\
  APNet-C$_{3}$ & 90.3 &96.1 & \bf 98.8 &  81.3 & 90.2 & \bf 95.8& 63.1& 82.8 & 91.2 \\
  \hline
  Stacked Attention (S) & 87.9 &95.5 & 98.5 & 77.6 & 88.6 & 94.7 &  56.1&78.7 & 89.0  \\
  APNet-S$_{1}$ & 88.0 &95.7 & 98.5 &  77.5 &88.8 & 94.7 & 55.8& 78.2 & 88.7   \\
  APNet-S$_{2}$ & 89.0 &\bf 96.1 & \bf 98.7 &  \bf 78.8 &\bf 89.3 & \bf 95.0 & 58.9& \bf 80.8 & 89.7 \\
  APNet-S$_{3}$ &\bf 89.3 & \bf96.1 & 98.6 & \bf78.8 & 89.2 & 94.8 & \bf 59.0& 80.7 & \bf 89.8  \\     
  \hline 
 \end{tabular}
\end{table*}

\subsection{Datasets and Experimental Settings} 
\textbf{Datasets:} 
We conduct extensive experiments on five widely used ReID datasets: Market-1501, DukeMTMC-reID, CUHK03, MSMT17 and Occluted-DukeMTMC. The detailed information of the datasets are shown in Table~\ref{tab:info_datasets}.. 
\subsubsection{Market-1501}
The Market-1501 dataset contains 751 identities with 12936 images for training and 750 identities with 19732 images for testing. All the images are captured by five high-resolution cameras and one low-resolution camera in a university. Deformable Part Model (DPM) is used as the pedestrian detector for the dataset. The author provided two kinds of query methods, and we follow the single-query method in this work.
\subsubsection{DukeMTMC-reID}
The DukeMTMC-reID dataset is a subset of the DukeMTMC dataset which contains 1,404 identities. 702 identities with 16522 images are selected as the training set and the remaining 702 identities with 19889 images are the testing set. The images are captured by 8 high-resolution cameras in Duke Univerisity and each identity is guaranteed to be observed by two cameras.
\subsubsection{CUHK03}
We conducted experiments on both versions of person boxes of the CUHK03 benchmark: manually labeled and auto-detected with a pedestrian detector. We chose the CUHK03-NP split in ~\cite{zhong2017re}, which selects 767 identities for training and the other 700 ones for testing. Compared with the 1367/100 split, the CUHK03-NP split is more realistic and challenging.
\subsubsection{MSMT17}
The MSMT17 dataset is the largest re-ID dataset, which contains 126,441 images of 4,101 identities captured by 15 cameras. In practice, 32,621 images of 1,041 identities are used for training and 93,820 images of 3,060 identities are used for testing. The dataset is recorded in 4 days with different weather conditions in a month using 12 outdoor cameras and three indoor cameras.

\textbf{Evaluation Metrics:} For all four datasets, we shared the same experiment settings with the standard person ReID experimental setups. We evaluated the ReID accuracy on four datasets by the cumulative matching characteristic (CMC) curve and mean Average Precision (mAP). CMC shows the ReID accuracy by counting the query identities among the top N results. The mAP score calculates the area under the precision-recall curve, which reflects the overall re-identification accuracy rather than only counting top N true matching. Note that, for all experiments, we directly calculate the distance with Euclidean distance, and do not employ the Re-ranking~\cite{zhong2017re} tricks.

\subsection{Implementation Details}
We adopt ResNet50~\cite{he2016deep} pre-trained on ImageNet~\cite{deng2009imagenet} as our backbone network for the experiments. 
The stride of the last residual block is set to 1 instead of the original $stride=2$ for a larger receptive region. We implement our APNet on both channel-wise and spatial attentions. The proposed attention pyramid is added after all four residual blocks. During training, three data augmentation methods including random cropping, horizontal flipping, and erasing are considered. The margin of triplet loss and the label smoothing regularization rate were set as 0.3 and 0.1, respectively. 
For the channel-wise attention pyramid, we follow the settings in~\cite{chen2019self} for a fair comparison. We use $384\times192$ input image size, and train the model 160 epochs with the Adam optimizer whose initial learning rate is 0.0004 and is divided by 10 every 40 epoch. 
For the spatial attention pyramid, we adopt the settings in~\cite{zhang2020relation} whose input images are resized to $256 \times 128$. We found that the multi-part trick in MGN~\cite{wang2018learning} is very effective
on the CUHK03 dataset but limited improvement on other datasets. Therefore, we only apply this multi-part trick on CUHK03.
For all experiments, we randomly select 20 persons in a batch for training, where each person has 4 images. 
During the evaluation, we use the $2048$ dimension features after the BN bottleneck for person matching. We employ the cosine distance as the evaluation metric to calculate the similarity of two images and use the average feature between the original testing image and the horizontally flipped one. All the experiments are conducted with PyTorch 1.7 with two Nvidia 2080 Ti GPUs.

\subsection{Ablation Studies}
To explore the effectiveness of each component of APNet and different hyper-parameters, we conducted comprehensive ablation studies on both spatial and channel-wise APNet. As shown in Table~\ref{tab: ablation}, We compare both spatial attention pyramid network (APNet-S) and channel-wise attention pyramid network (APNet-C) in different pyramid levels and multiple baselines including APNet$_{0}$, Stacked attention, and original SE-ResNet~\cite{hu2018squeeze}. APNet-S$_{3}$ denotes the network with 3-level spatial attention pyramid $\mathcal{P} = \{\mathcal{P}_1,\mathcal{P}_2,\mathcal{P}_3 \} $. APNet$_{0}$ is the baseline network without any attention model. Stacked attention denotes stacking same attentions without splitting as $\mathcal{P} = \{\mathcal{P}_1,\mathcal{P}_1,\mathcal{P}_1\} $. 

\subsubsection{Attention Pyramid Method vs. Baseline}
We first examined the effectiveness of the proposed attention pyramid method. Table~\ref{tab: ablation} shows the comparison of our APNet method with the baseline network without attention named as APNet$_{0}$. We observed that both APNet-S and APNet-C consistently obtain significant improvement over the baseline network. We achieve $+2.7\% / +1.2\%$ mAP/Rank1 performance improvement on the Market-1501 dataset and $+4.3\% / +2.0\%$ for DukeMTMC-reID. On the large scale MSMT17 dataset, we still outperformed $+12.3\% / +8.4\%$ and $+7.8\% / +5.4\%$ mAP/Rank1 by channel-wise and spatial APNet.

\subsubsection{Attention Pyramid vs. Stacked Attention}
To illustrate the superiority of the pyramid structure for the attention model, we designed a stacking scheme to prove that the naive linear attention stacking method is not helping the network to learn a more discriminative feature map. Specifically, instead of following our proposed "split-attend-merge-stack" pipeline introduced in Fig \ref{fig:method}, we just send the input feature map into three attention layers with the same architecture consecutively. The only difference between this stacked attention and our attention pyramid is the "split" mechanism. As shown in Table~\ref{tab: ablation}, we compared our APNet and stacked attention model on both spatial attention and channel-wise attention. With the 3 level attentions, our APNet-C$_{3}$ outperforms the stacked attention by a large margin on all four datasets, including $+0.9\% / +0.8\%$ mAP/Rank1 on Market-1501, $+2.6\% / +1.8\%$ mAP/Rank1 on DukeMTMC-reID, and $+7.9\% / +7.5\%$ mAP/Rank1 on MSMT17. The improvement is also significant for the spatial attention pyramid APNet-S. It demonstrates the effectiveness of the proposed "split" mechanism. Although more attention layers and parameters are used, the performance of stacked attention is even lower than the original single attention layer APNet-C$_{1}$ or APNet-S$_{1}$. It proves that more attention layers and parameters with the higher computational cost are not key factors for the network to learn a more robust feature representation, and directly proves the superiority of our proposed attention pyramid method is achieved by the effective pyramid structure but not more parameters.

\subsubsection{Influence of Pyramid Level}
In addition to the ``split'' mechanism, we also evaluated the effectiveness of the ``stack'' principle. For this goal, we compared the APNet with different attention map stacking schemes. Taking channel-wise attention pyramid network as an example, APNet-C$_{1}$ only has a global attention map $\{\mathcal{P}_1 \} $, while APNet-C$_{2}$ and APNet-C$_{3}$ respectively stack 2 and 3 attentions as $\{\mathcal{P}_1 \} $, $\{\mathcal{P}_1,\mathcal{P}_2 \} $, and $\{\mathcal{P}_1,\mathcal{P}_2,\mathcal{P}_3 \} $.  From Table~\ref{tab: ablation}, we can observe that the performance obviously increases when stacking fine-grained attention on the original global attention. On the Market-1501 and DukeMTMC-reID dataset, APNet-C$_{2}$ outperforms APNet-C$_{1}$ with $+0.9\% / +0.7\%$ and $+2.5\% / +1.8\%$ mAP/Rank1, while this improvement is larger for MSMT17 as $+4.0\% / +1.8\%$. When the attention pyramid goes deeper, the increase turns slow, which indicates adding the levels of the attention pyramid is rarely helpful when the level of the pyramid is enough. It is because the resolution is too limited to discover the discriminative semantic clues in deeper layers. Due to the trivial improvement and computational cost increment of the deeper attention pyramid, we choose the low-level attention pyramid for the following experiments.

\subsubsection{Channel-wise Attention vs. Spatial Attention}
We implement our attention pyramid network with both spatial and channel-wise attention. In both Market-1501 and DukeMTMC-reID datasets, we get very similar results with these two attention modules. However, we observe that our method with channel-wise attention is better than spatial attention on the MSMT17 dataset. For channel-wise attention, the Rank-1 and mAP accuracy are $3.7\% / 3.7\%$, $4.6\% / 2.9\%$ and $4.1\% / 2.1\%$ greater than spatial attention for APNet at pyramid level 1, level 2 and level 3, respectively. It might be due to the larger spatial variances with indoor and outdoor images, which causes the inaccurate spatial representation for the model. 

\begin{figure}[t]
	\centering
	\includegraphics[width=1.0\linewidth]{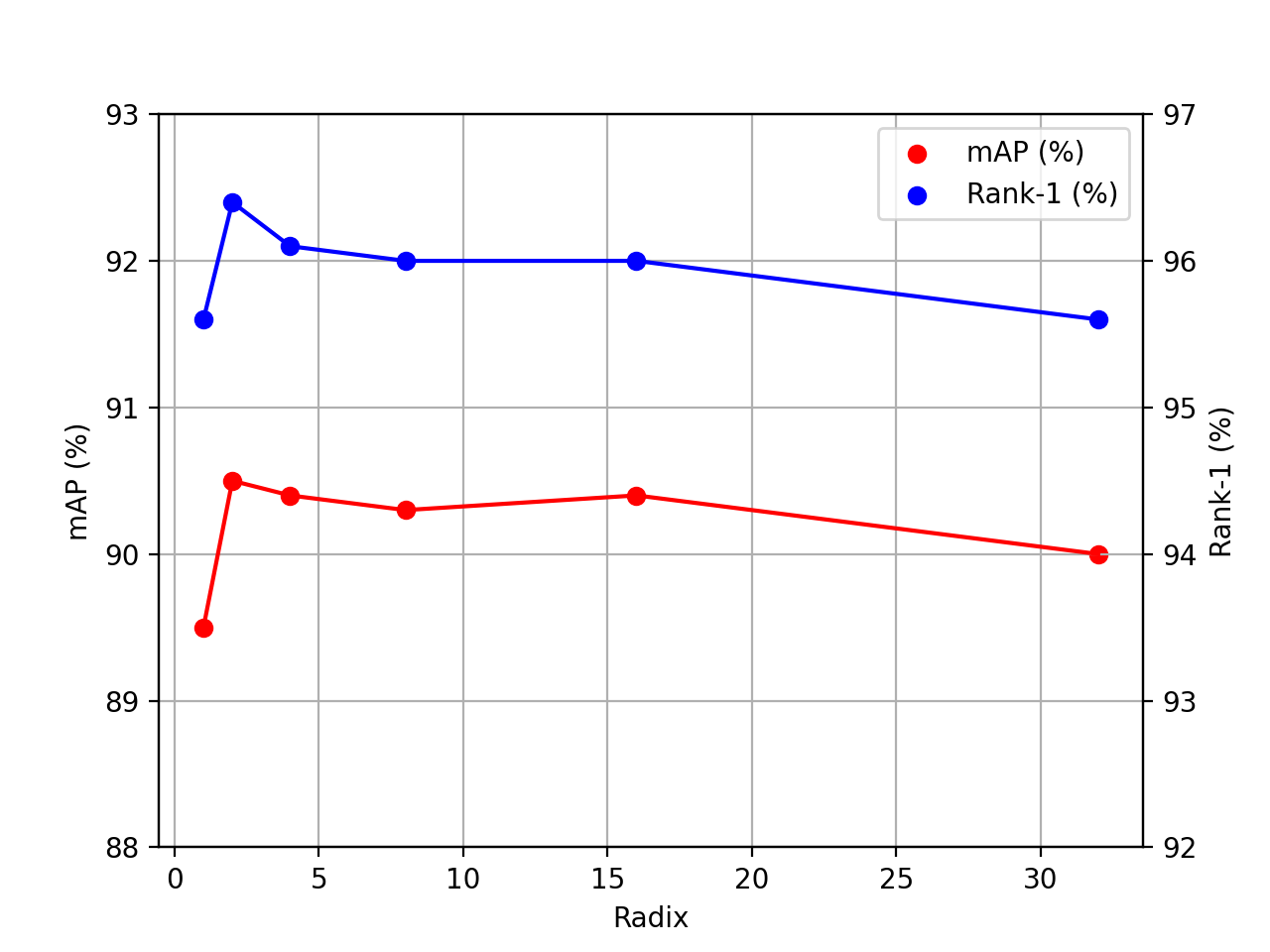}
	\caption{The mAP (red) and Rank-1 accuracy (blue) with different radix for APNet with level 2 pyramid on the Market-1501 dataset.}
	\label{fig:ablation_radix}
	\vspace{-0.2cm}
\end{figure}

\begin{table}[t]
    \centering
    \normalsize
    \caption{GFLOPS of different pyramid levels for APNet-C, SE-ResNet, Pyramid ReID and SCSN.}
    \label{tab: flop}
    \small
    \begin{tabular}{c | c }
    \hline
    \bf Methods &  \bf GFLOPS \\
    \hline
    \bf APNet-C$_0$ & 6.23  \\
    \bf APNet-C$_1$ & 6.23  \\
    \bf APNet-C$_2$ & 6.24  \\
    \bf APNet-C$_3$ & 6.25  \\
    \bf SE-ResNet~\cite{hu2018squeeze} & 6.24  \\
    \bf Pyramid ReID~\cite{zheng2019pyramidal} & 9.96 \\
    \bf SCSN~\cite{chen2020salience} & 7.24 \\
    \hline
\end{tabular}
\vspace{-0.2cm}
\end{table}

\subsubsection{Influence of Split Radix}
Beyond pyramid levels, we also conducted experiments to analyze the radix of the split process. The radix indicates the speed of scale reduction in our attention pyramid. In this experiment, we use attention pyramid networks at level 2 as the baseline model and apply different radixes for comparison. Specifically, we choose split radix as $r=\{ 1, 2, 4, 8, 16, 32 \} $ to respectively build our APNet, where radix $R-1$ denotes no splitting. The experiment results are shown in Fig\ref{fig:ablation_radix}. We can observe obvious maxima of both mAP and Rank-1 accuracy at radix equals to 2, and both accuracy slightly drops when radix increases. The phenomenon that performance drops while radix increases indicates the lower granularity is detrimental to discover the discriminative semantic clues. Such phenomenon is consistent with the observation that the improvement turns to slow with a deeper attention pyramid. It motivates us to maintain an adequate scale in the splitting process.

\begin{table*}[h]
\centering
\normalsize
\caption{Comparing with the state-of-the-art methods on the Market-1501, DukeMTMC-reID, MSMT17, and CHUK03 datasets. We apply the CUHK03-NP version for evaluation, where (d/l) denotes the performance on the detected and labeled data. $^* \ denotes\  the\   result\  is\  reproduced\  by\  us.$}
\renewcommand\tabcolsep{4pt}
\label{tab:sota_result}
\small
\begin{tabular}{c c c ccccccccccc}
  \hline
  {\multirow{2}{*}{\bf Method}}&
  {\multirow{2}{*}{\bf Publication}}&
  {\multirow{2}{*}{\bf Backbone}}&
  
  \multicolumn{3}{c}{\bf Market-1501}&
  \multicolumn{3}{c}{\bf DukeMTMC-reID}&
  \multicolumn{3}{c}{\bf MSMT17}&
    \multicolumn{2}{c}{\bf {CUHK03 (d/l)}}\\
  \cline{4-14}
   & & &\textbf{mAP}& \textbf{R-1}& \textbf{R-5}&\textbf{mAP}& \textbf{R-1}& \textbf{R-5}&\textbf{mAP}& \textbf{R-1}& \textbf{R-5}&   \textbf{{mAP}}& \textbf{{R-1}}\\
    \hline
  PCB+RPP~\cite{sun2018beyond}& ECCV'18&ResNet-50& 81.6& 93.8& 97.5& 69.2& 83.3& - & 40.4 & 68.2 & 81.6 & {57.5/-} & {63.7/-} \\
  MGN~\cite{wang2018learning}& ACMMM'18&ResNet-50& 86.9& 95.6& - & 78.4& 88.7& -& -& -& - & {66.0/67.4} & {66.8/ 68.0}\\
  VMP~\cite{sun2019perceive}& CVPR'19&ResNet-50& 80.8& 93.0& 97.8& 72.6& 83.6& 91.7& -& -& - & {-} & {-} \\
  DG-Net~\cite{zheng2019joint}& CVPR'19&ResNet-50& 86.0& 94.8& - & 74.8& 86.6& -& 52.3 & 77.2 & - & {61.1/-} & {65.6/-} \\
  BoT\cite{luo2019bag} & CVPRW'19 &ResNet-50 & 85.9 & 94.5 & - & 76.4 & 86.4 & - & -& -& - & {-} & {-}\\
  DSA~\cite{zhang2019densely} & CVPR'19 &ResNet-50& 87.6& 95.7& - & 74.3& 86.2& -& -& -& - & {73.1/75.2} & {78.2/78.9} \\
  Pyramid~\cite{zheng2019pyramidal} & CVPR'19 &ResNet-101 & 88.2 &95.7& 98.4 & 79.0 &89.0& -& -& -& - & {74.8/76.9} & {78.9/78.9} \\
  IANet\cite{hou2019interaction} &CVPR'19&ResNet-50 & 83.1 & 94.4 & - & 73.4 & 87.1 & - & 46.8 & 75.5 & 85.5 & {-} & {-} \\
  OSNet\cite{zhou2019omni} & ICCV'19 &OSNet & 84.9 & 94.8 & - & 73.5 & 88.6 & - & 52.9 & 78.7 & - & {67.8/-} & {72.3/-} \\
  SNR~\cite{jin2020style} & CVPR'20 & ResNet-50 & 84.7 &94.4 & - &  72.9 & 84.4 &- & - &- & -  & {-} & {-} \\
  ISP\cite{zhu2020identity} & ECCV'20 &HRNet-W32 & 88.6 & 95.3 & 98.6 & 80.0 & 89.6 & 95.5 & -& -& -& {71.4/74.1} & {75.2/76.5} \\ 
  CBN\cite{zhuang2020rethinking}  & ECCV'20 & ResNet-50 & 83.6 & 94.3 & 97.9 & 70.1 & 84.8 & 92.5& -& -& - & {-}& {-} \\
  {DPD~\cite{martinel2020deep}} & {TIP'20} & {ResNet-50} & {87.6} & {95.8} & {98.0} & {78.3} & {88.4} & {94.7} & {-} &{-} &{-} & {68.5/73.3} & {70.2/76.0}  \\
  {DPD-101~\cite{martinel2020deep}} & {TIP'20} & {ResNet-101} & {88.2} & {95.9} & {98.6} & {80.2} & {89.4} & {95.3} & {-} & {-} &{-} & {74.9/77.5} & {78.2/79.6} \\
  {CBDB-Net~\cite{tan2021incomplete}} & {TCSVT'21} & {ResNet-50} & {85.0} & {94.4} & {-} & {74.3} & {87.7} & {-} & {-} & {-} & {-} & {72.8/76.6} & {75.4/77.8} \\
  \hline
  MG-CAM\cite{song2018mask} & CVPR'18 &ResNet-50 & 74.3 & 83.8 & - & -& -& -& -& -& - & {38.6/41.0} & {41.7/44.4} \\
  HA-CNN~\cite{li2018harmonious}& CVPR'18 &HA-CNN& 75.7& 91.2& -& 63.8 & 80.5 & -& -& -& - & {38.6/41.0} & {41.7/44.4}\\
  DuATM\cite{si2018dual}& CVPR'18 &DenseNet & 76.6 &91.4 & 97.1 & 64.6 &81.8& 90.2& -& -& - & {-}& {-}\\
  SPReID~\cite{kalayeh2018human}& CVPR'18&  ResNet-152 & 83.4 & 93.7 & 97.6& -& -& -& -& -& - & {-}&{-} \\
  Mancs~\cite{Wang_2018_ECCV}& ECCV'18&  ResNet-50 & 82.3 & 93.1 & 97.6& 71.8& 84.9& -& -& -& - & {60.5/63.9} & {65.5/69.0} \\
  AAnet~\cite{tay2019aanet} & CVPR'19 & ResNet-50& 82.5& 93.9& -&72.6& 86.4& -& -& -& - & {-}& {-}\\
  SCAL-S\cite{chen2019self} & ICCV'19 & ResNet-50 & 88.9& 95.4& 98.5&79.6& 89.0& 95.1& -& -& - & {68.2/71.5} & {70.4/74.1} \\
  SCAL-C\cite{chen2019self} & ICCV'19 & ResNet-50 & 89.3& 95.8& 98.7&79.1& 88.9& 95.2& -& -& - & {68.6/72.3} & {71.1/74.8} \\
  CAMA~\cite{yang2019towards} & ICCV'19&ResNet-50 & 84.5 & 94.7 & 98.1 & -& -& -& -& -& - & {64.2/66.5} & {66.6/70.1}\\
  BAT-net\cite{fang2019bilinear} & ICCV'19&ResNet-50 & 85.5 & 94.1 & 98.2 & 77.3 & 87.7 & 94.7& 56.8 & 79.5& 89.1 & {73.2/76.1} & {76.2/78.6} \\
  MHN\cite{chen2019mixed} & ICCV'19&ResNet-50 & 85 & 95.1 & 98.1&77.2 & 89.1 & 94.6 & -& -& - & {65.4/72.4} & {71.7/77.2} \\
  ABD-Net\cite{chen2019abd} &ICCV'19&ResNet-50 & 88.3 & 95.6 & -&78.6& 89.0 & - & 60.8 & 82.3 & 90.6 & {-}& {-}\\
  {LAG-Net\cite{gong2021lag}} & {TMM'20}& {ResNet-50} & {89.5} & {95.6} & {98.3} & {\textbf{81.6}} & {\bf 90.4} & {\bf 96.0} & {-} & {-} & {-} & {79.1/82.4} & {82.2/85.1} \\
  SCSN\cite{chen2020salience} & CVPR'20 &ResNet-50 & 88.5 & 95.7 & -&79.0 & 90.1 & -& 58.0 & 83.0 & 91.2 & {80.2/83.3} & {\textbf{84.1}/86.3}\\
  RGA-SC\cite{zhang2020relation} &CVPR'20 & ResNet-50 & 88.4 & 96.1 & -& -& -& -& 57.5 & 80.3& - & {74.5/77.4} & {79.6/81.1}  \\
  $^*$RGA-S\cite{zhang2020relation} &CVPR'20 & ResNet-50  & 88.0 &95.7 & 98.5 &  77.5 &88.8 & 94.7 & 55.8& 78.2 & 88.7 &  {72.7/75.6} & {78.1/79.1} \\
  PISNet\cite{zhao2020not} & ECCV'20 & ResNet-50 & 87.1 & 95.6 & - & 78.7 & 88.8 & - & -& -& -& {-}& {-}\\
  \hline
  APNet-S& & ResNet-50& 89.0 & 96.1& 98.7 & 78.8& 89.3& 95.0 & 59.0 & 80.8 & 89.8 &  {78.1/81.1} & {80.9/83.5} \\
  APNet-C& &ResNet-50& \textbf{90.5}& \bf 96.2 & \textbf{98.8} & 81.5 &\bf 90.4 & 95.6 &\textbf{63.5}&\textbf{83.7} & \textbf{91.7} &  {\textbf{81.5/85.3} }& {83.0/\textbf{87.4}} \\
  \hline
\end{tabular}
\vspace{-0.1cm}
\end{table*}

\subsubsection{Analysis of computational cost}

A core advantage of our APNet is the more efficient computing than other pyramid structures. To evaluate the efficiency of the proposed attention pyramid, we compare the computational cost of APNet and other methods such as SE-ResNet~\cite{hu2018squeeze}, Pyramid ReID~\cite{zheng2019pyramidal} and SCSN~\cite{chen2020salience}. As shown in Table~\ref{tab: flop}, we summarize the Giga floating-point operations per second (GFLOPS) to represent the computational cost. Specifically, we use a single image resized to $224\times224$ as the input and calculate the GFLOPS by mainstream tool PyTorch-OpCounter\footnote{https://github.com/Lyken17/pytorch-OpCounter} for all experiments. The stride of the last residual block of ResNet in every tested method is set to 1.  As shown in Table~\ref{tab: ablation} and Table~\ref{tab: flop}, our APNet achieves higher accuracy with comparable computational cost compares with SE-ResNet~\cite{hu2018squeeze},. Specifically, APNet-C  outperforms SE-ResNet by $+1.6\% / +0.7\%$, $+3.4\% / +1.1\%$ and $+5.0\% / +2.6\%$ mAP/Rank1 respectively on three datasets with similar computational cost. It is because we only apply the attention pyramid on the top of each CNN block but not every convolution layer. We also compare our APNet with other feature pyramid structures such as Pyramid ReID~\cite{zheng2019pyramidal}. The comparison results show our APNet saves almost $40\%$ computational cost than Pyramid ReID, by reducing multiple convolution branches and using attention instead. We did not compare APNet with FPN~\cite{lin2017feature} or FPA~\cite{li2018pyramid}, since these methods focus on detection or segmentation tasks whose GFLOPS is significantly larger than APNet for the recognition task. Compared with the cascaded attention model SCSN~\cite{chen2020salience}, our attention pyramid is more efficient by $14.0\%$ GFLOPS. It is mainly because APNet needs no extra salience selection and information aggregation modules.

\subsubsection{  Generalizability for different backbones}
In most experiments of this paper, we apply the ``ResNet50'' as the backbone for a fair comparison with others. It is because most methods apply the ``ResNet50'' as their backbones in the field of person ReID. To evaluate the generalizability of our method for different backbones, we apply APNet into different deep learning architectures including ``ResNet101''~\cite{he2016deep}, ``DesNet169''~\cite{huang2017densely}, and ``Inception-V3''~\cite{szegedy2016rethinking}. In Table~\ref{ablation_backbones}, we compare the experimental performance of these deep learning architectures with/without our APNet on the Market-1501~\cite{zheng2015scalable} dataset. In this experiment, we first evaluate different deep learning architectures as baselines, and apply our APNet-C in these backbones. For both baseline backbone and our method, we employ the same loss functions, hyper-parameters, and training settings. We observe that our APNet can achieve consistent improvement on all backbone architectures. For ResNet50, our APNet achieved 2.7\% improvement on mAP score and 1.2\% improvement on Rank-1 accuracy. APnet also obtained excellent performance with ResNet101 backbone with 91.2\% mAP score and 90.4\% Rank-1 accuracy. It indicates that APNet is effective for different backbone scales. APnet also improved the Inception-V3 and DesNet169 in a large margin, i.e., 1.7\% mAP score and 0.9\% Rank-1 accuracy for Inception-V3 and 1.5\% mAP score and 1.1\% Rank-1 accuracy for DesNet169. It demonstrates that APNet is effective for different architectures.

\begin{table}[t]
\caption{ Ablation studies of different backbones on the Market-1501 dataset.}
\label{ablation_backbones}
\renewcommand\tabcolsep{9.0pt}
\begin{center}
\begin{tabular}{c|cccc}
\toprule
\bf Rank@R  &\bf mAP&\bf R-1 &\bf R=5 \\
\midrule
ResNet50  & 87.8 & 95.0 & 98.6 \\
ResNet50 + APNet-C &90.5 & 96.2 & 98.8 \\
\midrule
ResNet101  & 89.6 & 95.3& 98.5 \\
ResNet101 + APNet-C & 91.2 & 96.4 & 98.8 \\
\midrule
Inception-V3 & 86.3 & 95.1 & 98.2 \\
Inception-V3 + APNet-C & 88.0 & 96.0 & 98.5 \\
\midrule
{DesNet169}  & 89.0 & 95.1 & 98.7 \\
DesNet169 + APNet-C& 90.5 & 96.2 & 98.9 \\
\bottomrule
\end{tabular}
\end{center}
\end{table}

\subsection{Comparison to State-of-the-Art Methods}
To show the effectiveness of our proposed method, We compared the mAP, Rank-1, and Rank-5 accuracy of APNet with several state-of-the-art ReID methods on the popular Market-1501, DukeMTMC-reID, CUHK03, and MSMT17 datasets. Table \ref{tab:sota_result} summarizes the results of the comparison for different methods on all four datasets. We introduce the results without attention model on the top of the Table and report the performance of attention-based methods in the second part. Finally, we show both spatial and channel-wise APNet at the bottom of the Table. We observe that APNet achieves superior performance over all listed methods on four benchmarks, which illustrates the effectiveness of our method.

On the \textbf{Market-1501} dataset, our APNet achieves the state-of-the-art performance with the efficient attention pyramid mechanism. RGA-SC~\cite{zhang2020relation} is the current state-of-the-art method that uses a relation-aware global attention, and we improve the performance by 1.1\% and 0.1\% on mAP and Rank-1 accuracy. We also reproduce the result of RGA-S which only uses the spatial attention as our method. The result shows our APNet-S with pyramid structure helps the RGA-S improves $1.0\%$ and $0.4\%$ on mAP and Rank-1.
Pyramid ReID~\cite{zheng2019pyramidal} crops the input image into pieces with different scales and captures multi-scale clues with multiple convolution branches, which achieves the best performance without attention. We outperform Pyramid ReID by $2.3\%$, $0.5\%$ on mAP and Rank-1, respectively. Beyond performance, we also save $37.4\%$ computational cost than Pyramid ReID by the efficient attention-based pyramid structure.

On the \textbf{DukeMTMC-reID} dataset, APNet outperforms the second best method ISP~\cite{zhu2020identity} by 1.5\%/0.5\% on the mAP/Rank-1 performance. SCSN~\cite{chen2020salience} cascades multiple refine stages including an attention model and salience selection model, which obtains the state-of-the-art Rank-1 performance. However, this cascaded attention structure needs a complex mechanism with large computational cost to avoid information duplication such as the salience selection model. Our APNet achieves the same Rank-1 accuracy and a large improvement of $+2.5\%$ for mAP score with $14.0\%$ less computation.

On the \textbf{CUHK03} dataset, our method achieved performance improvement on both detected and labeled settings. SCSN\cite{chen2020salience} method achieves the state-of-the-art performance on the CUHK03 dataset which outperforms other methods by a large margin. Compared with SCSN\cite{chen2020salience}, our APNet obtains better performance. Specifically, we obtain 85.3\% mAP score and 87.4\% Rank-1 accuracy on the manually labeled data, and obtain 81.5\% mAP score and 83.0\% Rank-1 accuracy on the auto-detected one. We find that the multi-part trick in MGN~\cite{wang2018learning} is very effective on the CUHK03 dataset. The multi-part trick adds a new local branch which splits the feature map into two parts and learns the local features. Specifically, the feature maps are split from the third residual block along the height dimension. In the inference, we connected the original global feature with local features for final matching. However, the effectiveness of this trick is limited when we add the scale of datasets (e.g., MSMT17~\cite{wei2018person} or Market-1501~\cite{zheng2015scalable}). Thus, we only use the mutli-part trick on the CUHK03 dataset, but not on other datasets.

\textbf{MSMT17} is the large-scale person ReID dataset with both indoor and outdoor images. As shown in Table~\ref{tab:sota_result}, the attention mechanism shows great effectiveness on this challenging dataset due to discovering discriminative clues. While our APNet further obtains the improvement than the best published method ABD-Net\cite{chen2019abd} by a large margin. We obtain $63.5\%/83.7\%$ Rank1/mAP, which outperforms ABD-Net\cite{chen2019abd} with $2.7\%/1.4\%$ by the effective pyramid structure. 

\begin{table}[t]
	\centering
	\normalsize
	\caption{The comparisons with the state-of-the-art methods on the Occluded-DukeMTMC dataset.}
	\label{tab:occ_duke_result}
	 \small
	\begin{tabular}{c|cccc}
		\hline
		\textbf{Method} & \textbf{mAP}& \textbf{R-1}& \textbf{R-5}& \textbf{R-10}\\
		\hline
		HA-CNN~\cite{li2018harmonious}& 26.0& 34.4& 51.9& 59.4\\
		Adver Occluded~\cite{huang2018adversarially}& 32.2& 44.5& - -& - - \\
		PCB~\cite{sun2018beyond}& 42.6& 37.3& 57.7& 62.9\\
		Part Bilinear~\cite{suh2018part}& -& 36.9& -& -\\
		FD-GAN~\cite{ge2018fd}& -& 40.8& -& -\\
		\hline
		DSR~\cite{he2018deep}& 30.4& 40.8& 58.2& 65.2\\
		SFR~\cite{he2018recognizing}& 32.0& 42.3& 60.3& 67.3\\
		Ad-Occluted\cite{huang2018adversarially} & 32.2 & 44.5 & - & - \\
		PGFA\cite{miao2019PGFA} & 37.3 & 51.4 & - & - \\
		HOReID\cite{wang2020high} &43.8&55.1& -  & -\\	
		\hline
		\bf APNet-C$_{1}$& 46.6& 53.9 & 68.6& 74.1\\
		\bf APNet-C$_{2}$& \bf 54.1& \bf 62.2 & \bf 76.3& \bf 81.5\\
		\hline
	\end{tabular}
\end{table}

\begin{table}[t]
	\centering
	\normalsize
	\caption{Cross-domain evaluation on Market-1501 and DukeMTMC-ReID datasets. We use Market $\rightarrow$ Duke to represent that the model is trained on the Market-1501 dataset and tested on the DukeMTMC-ReID dataset, and vice versa. }
	\label{tab: cross-domain}
	\renewcommand\tabcolsep{3pt}
	 \small
	\begin{tabular}{l|*{4}{c}|*{4}{c}}
		\hline
		\multirow{2}*{\bf Method} & \multicolumn{4}{c|}{Market$\rightarrow$Duke} &\multicolumn{4}{c}{Duke$\rightarrow$Market}\\
		\cline{2-9}
		& {\bf mAP}& {\bf R-1}&{\bf R-5}&{\bf R-10}&{\bf mAP}&{\bf R-1}&{\bf R-5}&{\bf R-10}\\
		\hline
		Baseline  & 15.6 & 29.1&43.4&50.1& 19.3& 44.4&61.1&66.7\\
		SCAL-C & 16.4 & 28.6&-&-& 23.8&\bf 51.7&-&-\\	
		APNet-C$_{1}$  & 16.6 & 30.1&43.9&50.0& 23.0& 50.4&65.2&71.8\\
		APNet-C$_{2}$ & 21.3& 35.9& 50.1& 56.9&\bf 24.0&51.0&66.4& 72.3\\ 
		APNet-C$_{3}$ & \bf 22.8&\bf 37.7&\bf 52.4&\bf 59.0&23.7&50.9&\bf 66.6&\bf 72.6\\ 
		\hline  
	\end{tabular}
\end{table}

\begin{figure*}[t]
	\centering
	\includegraphics[width=0.99\linewidth]{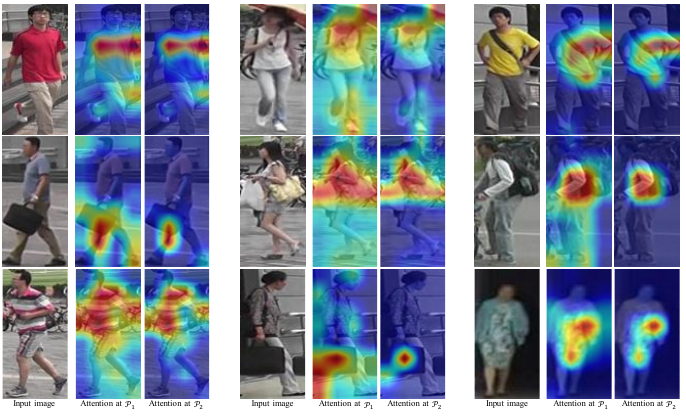}
	\caption{
	Visualizations of the attention map with different pyramid levels. We adopt the Grad-CAM~\cite{selvaraju2017grad} to visualize the learned attention maps of our attention pyramid. For each sample, from left to right, we show the input image, attention of the first level pyramid, attention of the second level pyramid. We can observe that attentions in different pyramid levels capture the salient clues of different scales. Best viewed in color.}
	\label{fig:vis}
\end{figure*}

\subsection{Robustness Analysis}
To further evaluate the robustness of APNet, we test our method on the Occluded-DukeMTMC dataset which contains occluded and corrupted inputs. Occluded-DukeMTMC is transformed from DukeMTMC-reID by re-splitting the size of each subset manually.
Original DukeMTMC-reID dataset includes 14\%, 15\%, and 10\% of occluded images in training, query, and gallery dataset, while re-split Occluded-DukeMTMC contains 9\%, 100\%, and 10\% of occluded images in training, query, and gallery dataset,  respectively. As a consequence, at least one of the features extracted by images will from occluded images and used in pairwise distance calculation at inference time. We follow the experimental settings in~\cite{miao2019PGFA} of the Occluded-DukeMTMC dataset. It is a direct metric to evaluate the robustness of the model trained on the original images for the testing of occluded images.

Recently, many methods~\cite{huang2018adversarially,miao2019PGFA,wang2020high,he2018deep,he2018recognizing} focus on the robustness of the ReID model for occlusion and achieve obvious success. We also test APNet on the Occluded-DukeMTMC dataset and compare it with other methods designed for the occlusion problem. Table~\ref{tab:occ_duke_result} shows the comparison between our APNet and the current state-of-the-art methods on this dataset. From the result, APNet achieves 54.2\% and 62.2\% in mAP and Rank-1 accuracy, respectively, which outperforms the current best model by a large margin as +10.3\% and +7.1\%. It demonstrates the attention model helps the network to focus on the salient region and avoid the corruption by occlusion. While the large improvement of the pyramid structure APNet-C$_{2}$ over the baseline attention model APNet-C$_{1}$ shows the pyramid structure enhances the attention model to capture the multi-scale salient clues to further prevent absorbing occluded feature representation.

\subsection{Generalization Ability Analysis}
In real-world applications, we always need to deploy the ReID model into unseen scenes. However, it requires extensive human labor to label an overwhelming amount of data for training models on new scenes. Thus, the generalization ability of the person ReID methods becomes a key factor for deploying the real-world application. To evaluate the generalization ability of our APNet, we conducted a cross-dataset evaluation to measure the generalization ability of the ReID model for unseen persons and scenes. Specifically, we trained the model on the Market-1501 dataset then tested it on DukeMTMC-reID, and vice versa. We conducted the experiments for APNet with different levels and the baseline model and compared it with other attention-based methods such as SCAL~\cite{chen2019self}. The results in Table~\ref{tab: cross-domain} show our APNet achieves +5.7\%/ +6.8\% mAP/Rank-1 performance over the baseline methods from Market to Duke, and obtain the improvement +4.6\%,+6.4\% mAP/Rank-1 from Duke to Market. It shows that the pyramid structure is effective to enhance the generalization ability of the attention model. Compared with SCAL~\cite{chen2019self}, our APNet achieves the comparable Duke$\rightarrow$Market performance and significantly improved Market$\rightarrow$Duke performance, which also shows the great potential of attention pyramid structure in terms of generalization ability.

\subsection{Qualitative Analysis}
To validate the effectiveness of our attention pyramid networks learning method, we used Grad-CAM~\cite{selvaraju2017grad} to visualize the attention map for qualitative analysis. The attention map is generated after each pyramid level, and we expect the attention mask to emphasizes more on the discriminative feature of the person at the deeper pyramid level. As shown in Fig.~\ref{fig:vis}, we can observe the salient feature of the person such as the handbag, umbrella, logo on the cloth, and shirt are highlighted. When the pyramid level goes deeper, we find the attention map is more concentrated on the salient part of the person and alleviate the common misalignment issue in image retrieval task.

\section{Conclusion}
In this paper, we have proposed simple yet effective attention pyramid networks (APNet) for the person re-identification task. To capture the salient clues with different scales, we proposed a ``split-attend-merge-stack'' principle to build the  attention pyramid. We split the feature maps into local parts and merge all learned local attentions as global attention. By stacking the attention modules with different granularities of splitting, we construct an attention pyramid to guide the fine-grained attention learning with coarse ones. We implement our APNet with both spatial and channel-wise attention modules to demonstrate it can be integrated into any attention model. By extensive experiments, we also demonstrate that APNet is more effective, efficient, and robust than other pyramid structures or attention models.

\section*{Acknowledge}
This work was supported in part by the National Key R$\&$D Program of China under Grant 2020AAA0105220, in part by the National Natural Science Foundation of China under Grant 61822603, Grant U1813218, and Grant U1713214, in part by a grant from the Beijing Academy of Artificial Intelligence (BAAI), and in part by a grant from the Institute for Guo Qiang, Tsinghua University.

{
\bibliographystyle{IEEEtran}
\bibliography{att_py}
}

\end{document}